\documentclass[10pt,twocolumn,letterpaper]{article}

\usepackage{graphicx}
\usepackage{cvpr}
\usepackage{stfloats}
\usepackage{multirow}
\usepackage{booktabs}
\usepackage{makecell}

\usepackage[pagebackref,breaklinks,colorlinks]{hyperref}
\usepackage[breaklinks,colorlinks]{hyperref}
\usepackage[capitalize]{cleveref}

\crefname{section}{Sec.}{Secs.}
\Crefname{section}{Section}{Sections}
\Crefname{table}{Table}{Tables}
\crefname{table}{Tab.}{Tabs.}

\title{PAMI: partition input and aggregate outputs for model interpretation}

\author{Wei Shi\\
Sun Yat-sen University\\
{\tt\small shiw58@mail2.sysu.edu.cn}
\and
Wentao Zhang\\
Sun Yat-sen University\\
{\tt\small zhangwt65@mail2.sysu.edu.cn}
\and
Weishi Zheng\\
Sun Yat-sen University\\
{\tt\small wszheng@ieee.org}
\and
Ruixuan Wang\\
Sun Yat-sen University\\
{\tt\small wangruix5@mail.sysu.edu.cn}
}

\begin{document}
\maketitle
\begin{abstract}
There is an increasing demand for interpretation of model predictions especially in high-risk applications. Various visualization approaches have been proposed to estimate the part of input which is relevant to a specific model prediction. However, most approaches require model structure and parameter details in order to obtain the visualization results, and in general much effort is required to adapt each approach to multiple types of tasks particularly when model backbone and input format change over tasks. In this study, a simple yet effective visualization framework called PAMI is proposed based on the observation that deep learning models often aggregate features from local regions for model predictions. The basic idea is to mask majority of the input and use the corresponding model output as the relative contribution of the preserved input part to the original model prediction. For each input, since only a set of model outputs are collected and aggregated, PAMI does not require any model details and can be applied to various prediction tasks with different model backbones and input formats. Extensive experiments on multiple tasks confirm the proposed method performs better than existing visualization approaches in more precisely finding class-specific input regions, and when applied to different model backbones and input formats. The source code will be released publicly. 

\end{abstract}

\section{Introduction}
Deep learning models have shown human-level performance in various machine learning tasks and started to be applied in real scenarios, such as face identification~\cite{he2018wasserstein,yan2019vargfacenet,kim2020groupface}, medical image analysis~\cite{chen2019med3d,li2021medical,sanderson2022fcn}, and language translation~\cite{wu2021r,xu2021bert,takase2021rethinking}. However, current deep learning models often lack interpretations for their decision making, which hinders the massive deployment of intelligent systems  particularly in high-risk applications like medical diagnosis and autonomous driving.  

To improve interpretation of model predictions, multiple visualization approaches have been proposed to localize input regions or components which are more relevant to the model prediction given any specific input to the model. For image classification task as an example, the class activation map (CAM) and its variants utilize the output (i.e., feature maps) of certain convolutional layer in the convolutional neural networks (CNNs) and their contribution weights to find the image regions which are responsible for the specific model prediction given any input image~\cite{zhou2016learning,selvaraju2017grad,chattopadhay2018grad,wang2020score,liu2022partial}, and the back-propagation approaches propagate the CNN output layer-by-layer to the input image space either based on gradient information (or its modified versions) at each layer~\cite{springenberg2015striving, selvaraju2017grad, chattopadhay2018grad, srinivas2019full, sundararajan2017axiomatic} or based on the relevance between input elements and output at each layer~\cite{binder2016layer, bach2015pixel, montavon2017explaining, iwana2019explaining}.
While the CAM-like approaches can only roughly localize relevant regions due to the lower resolution of feature maps, the back-propagation approaches often just find sparse and incomplete object regions relevant to the model prediction. Moreover, both types of approaches need either part of or the whole model structure and parameter details, which may be unavailable in some applications due to privacy or security concerns. When the model details are not available, the occlusion method may be utilized to roughly localize image regions relevant to the model prediction by occluding each local patch and checking the change in model output~\cite{zeiler2014visualizing,petsiuk2018rise}. However, the occlusion method often only localizes the most discriminative object part and misses the other parts which actually also contribute to the model prediction. LIME~\cite{ribeiro2016should} is another method without requiring model details for model interpretation by locally approximating the model decision surface for any specific input, but it can often estimate the contributions of a subset of input parts and require an optimization process for interpretation of a specific model prediction. Furthermore, most existing visualization approaches for interpretation of model predictions are developed for specific type of tasks (e.g., just for image classification), model backbone (e.g., for CNNs), and input format (e.g., just for image data). Substantial efforts are often required to adapt one visualization approach to various tasks (e.g., image caption) with different model backbones (e.g., Transformer backbone) or input formats (e.g., sequence of items). 

Different from existing visualization approaches, a simple yet effective visualization framework for interpretation of model predictions is proposed in this study. The proposed framework, called PAMI (`Partition input and Aggregate outputs for Model Interpretation'), is inspired by the observation that both humans and popular deep learning models extract and aggregate features of local regions for image understanding and decision making. 
Suppose a well-trained image classifier predicts an input image as a specific class. To find the relevant image regions and their contributions to the model prediction, the proposed framework first partitions the input into multiple parts, and then feeds only one part (with the remaining regions masked) to the model to obtain the corresponding output probability of the specific class. Aggregating the output probabilities over all the individual parts would result in an importance map representing the contribution of each input part to the original model prediction. In contrast to existing visualization approaches, the proposed PAMI framework does not require model structure and parameter details, can more likely find all possible input parts which are relevant to the model prediction, more precisely localize relevant parts, and work for various model backbones with different input formats. Such merits of the proposed PAMI framework has been confirmed by extensive experiments on multiple tasks with different model backbones and input formats. 

\section{Related Work}

In computer vision, post-hoc interpretation of deep learning models focuses on either understanding of model neurons (e.g.,
convlutional kernels, output elements) which is independent
of input information \cite{erhan2009visualizing, mahendran2015understanding, jason2015understanding, kim2018interpretability, bau2020understanding}, or understanding of a specific model prediction given an input image~\cite{lim2021building, petsiuk2018rise, elliott2021explaining, selvaraju2017grad, chattopadhay2018grad}. This study belongs to the latter one, i.e., trying to understand what information in the input causes the specific model prediction.

Multiple approaches have been proposed for understanding of model predictions, including the activation map approach~\cite{chattopadhay2018grad,selvaraju2017grad,zhou2016learning, jiang2021layercam}, the back-propagation approach~\cite{iwana2019explaining, zeiler2014visualizing, gu2018understanding, zeiler2014visualizing}, the perturbation approach~\cite{petsiuk2018rise,fong2017interpretable}, 
the local approximation  approach~\cite{ribeiro2016should, zhou2015object} and the optimization-based approach~\cite{lim2021building, elliott2021explaining}. 
The activation map approach often obtains a class-specific activation map with the weighted sum of all feature maps often at the last convolutional layer, and considers the regions with stronger activation relevant to the specific model output~\cite{chattopadhay2018grad,selvaraju2017grad,wang2020score,zhou2016learning}. 
Since the activation map is often much smaller than the input image, only approximate image regions corresponding to the stronger activation regions can be localized for interpretation of the specific model prediction.
Different from the activation approach which often works at higher layer of the deep learning model, the back-propagation approach tries to estimate the importance of each input pixel by propagating the specific model output layer-by-layer back to the input space. This can be obtained by calculating the gradient of the specific model output with respect to input elements at each layer~\cite{springenberg2015striving, selvaraju2017grad, chattopadhay2018grad, srinivas2019full, sundararajan2017axiomatic}, the input-relevant contribution of each kernel at each layer~\cite{kindermans2018learning}, or the relevance between output and each input element at each layer~\cite{binder2016layer, bach2015pixel, montavon2017explaining, iwana2019explaining}. The back-propagation approach considers each input pixel as an independent component and often only a subset of disconnected pixels in the relevant regions are estimated to be relevant to the model prediction. 

To find local image regions rather than disconnected pixels relevant to the model prediction, the perturbation approach has been proposed by perturbing local image regions somehow and checking the change in model output of the predicted class~\cite{fong2017interpretable, zeiler2014visualizing}. If certain perturbed local region causes large drop in the output, the local region in the input image is considered crucial to the original model prediction. Perturbation can be in the form of simply masking a local region by a constant pixel intensity~\cite{zeiler2014visualizing}, by neighboring image patches~\cite{zintgraf2017visualizing}, or by blurring the original region information with a constant value or smoothing operator~\cite{petsiuk2018rise, fong2017interpretable}. 
This approach often can find only the most discriminative part of the relevant regions which are responsible for the model prediction, because perturbing less-discriminative part of the relevant regions often does not cause much drop in model output. 
Besides the perturbation approach, the local approximation approach provides another way to estimate the contribution of each meaningful image region (e.g., object parts, background region) to the model prediction~\cite{ribeiro2016should, wang2020score, fong2017interpretable}. 
This approach assumes that the model decision surface is locally linear in the region-based feature space for any specific input, 
and therefore can be approximated with a linear model in the feature space. The weight parameters in the linear model can directly indicate the contribution of each meaningful image region to the original model output, thus obtaining image regions most relevant to the model prediction. 

While most of these visualization approaches to interpretation of model predictions were originally developed for image classification models, they have been extended or modified for other tasks~\cite{danilevsky2020survey, wallace2020interpreting} or other deep learning models~\cite{chefer2021transformer, li2019graph}. Besides these approaches,  prototype-based~\cite{nauta2021neural, rymarczyk2021protopshare} and and attention-based~\cite{obeso2022visual} approaches have also been proposed for model interpretation. Note that except the local approximation approach and part of the perturbation approach (e.g., the occlusion method~\cite{fong2017interpretable}, most approaches require at least part of the model structure and parameter details in order to find input parts which are relevant to the model prediction. 


\begin{figure*}[htbp]
  \centering
   \includegraphics[width=1\linewidth]{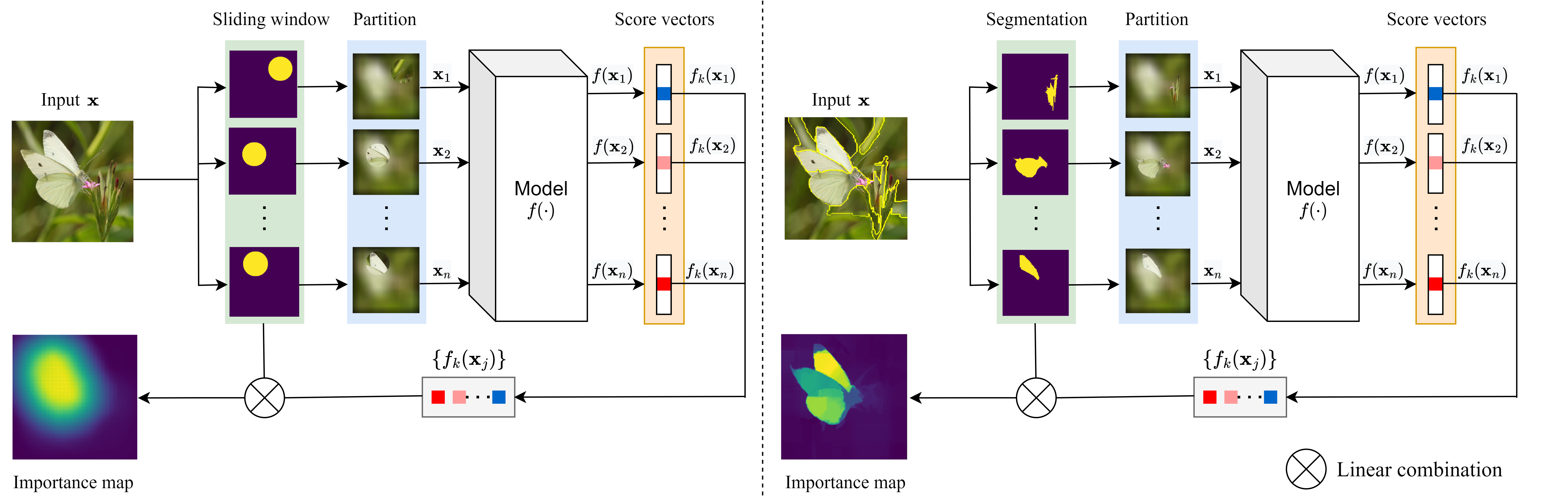}
   \caption{The proposed PAMI framework with the sliding window based (left half) or the pre-segmentation based (right half) input partition strategy. Each time only one local part of the input is preserved and the remaining parts are masked (blurred here).   
   } 
   \label{fig:pami}
\end{figure*}

\section{Method}


In this study, we aim to provide interpretation for model prediction given any specific input to a well-trained and fixed deep learning model. The interpretation is demonstrated by estimating relative contribution of each input part to the specific model prediction and correspondingly localizing input regions or elements which are relevant to the model prediction. It is worth noting that no model structure and parameter details are assumed to be known during the model interpretation process.

\subsection{Motivation}
Although humans can often instantly recognize objects in images, certain attention mechanism in human brain is likely involved in the process of object recognition~\cite{itti2001computational, petersen2012attention}. In other words, humans often need to implicitly or explicitly attend to local regions for image understanding and object recognition. While the detailed human attention mechanism is yet to be further explored, initial studies~\cite{desimone1995neural} suggest that most local parts of an object in an image help humans recognize the object, and  appearance of only an individual object part could help humans recall the corresponding class of the object. Consistent with the visual attention studies, recent exploration of convolutional neural networks (CNNs) shows that convolutional kernels even at higher convolutional layers (i.e., closer to the CNN output) often have smaller receptive fields than expected~\cite{luo2016understanding}. Considering that a global pooling is performed at the last convolutional layer in most CNN classifier models, it is widely accepted that CNN models largely depend on the collection of local image region features for image classification. For the other type of deep learning model backbone Transformer and its variants (e.g., ViT~\cite{dosovitskiy2020image}, Swin Transformer~\cite{liu2021swin}), since most items in the input sequence at each model layer correspond to components (e.g., words for a sentence input, image patches for an image input) of the original input, the final model prediction also largely depends on the collection of local features of the original input. With the above observation, we hypothesize that the model output response to each single component of the original input may directly imply the importance of the single input component for the specific model prediction.

\subsection{The proposed PAMI framework}
The proposed interpretation framework is demonstrated in Figure~\ref{fig:pami}. For image classification task as an example,
given a well-trained classifier model $f(\cdot)$ and any input image $\mathbf{x}$, denote by $f_c(\mathbf{x})$ the output prediction probability of the input image belonging to the $c$-th class.  Suppose the model predicts the input as the $k$-th class, i.e., $f_k(\mathbf{x})$ is the maximum over all the output probabilities. To interpret the classifier's prediction, the proposed framework first partitions the input into multiple either overlapped or non-overlapped parts (see the following subsections), and then with the $j$-th individual part preserved and all the remaining image regions masked, the output probability $f_k(\mathbf{x}_j)$ of the $k$-th class for the majority-masked input $\mathbf{x}_j$ is used to estimate the relative contribution of the preserved $j$-th image part to the original model prediction $f_k(\mathbf{x})$. By collecting and aggregating the output responses $f_k(\mathbf{x}_j)$'s over all the partitioned image parts, an importance map with the same spatial size as that of the input image $\mathbf{x}$ can be generated to represent the contribution of each input element (i.e., pixel here). Local image regions with correspondingly higher response values in the importance map are supposed to contribute more to the model prediction $f_k(\mathbf{x})$, thus providing visual evidence for the model prediction. It is worth noting that the interpretation framework can be applied to different tasks (e.g., image caption and sentiment analysis) with various input formats. 

\subsubsection{Input partition strategy I: sliding window}
One simple way to partition input is to apply the sliding window strategy with a pre-defined window size and sliding step size, where the window is in certain regular shape (e.g., circular or rectangular). In this way, the original input can be easily partitioned into multiple parts, and each part can be more or less overlapped by its neighboring parts determined by window size and sliding step size. For each partitioned part, all the remaining image regions will be masked somehow (e.g., by black pixels or blurred version of the original regions), and the output probability of the originally predicted class can be directly obtained with the majority-masked image as input.
Since each input element (e.g., pixel of an input image) could be covered by multiple partitioned parts, the contribution of each input element in the final importance map can be obtained by averaging the output probabilities of the originally predicted class over all the partitioned parts covering the input element. 

Note that the window size would affect the resolution level of the final importance map. Although smaller window would result in desired higher resolution, an image part with much smaller size (i.e., too small window) could contain little semantic information such that it becomes challenging for the framework to estimate the contribution of the smaller image part to a specific model prediction. In practice, users can choose one appropriate window size for interpretation of model prediction, or multiple window sizes for interpretation at multiple scales of image parts. 


\subsubsection{Input partition strategy II: pre-segmentation}
Another way to partition input is to pre-segment the input into multiple parts with certain segmentation strategy. When the input is an image, various unsupervised segmentation algorithms can be adopted for pre-segmentation of the input. In this study, super-pixel segmentation algorithms are used for input image partition~\cite{felzenszwalb2004efficient, ren2003learning, meyer1992color, bergh2012seeds}. With a particular super-pixel segmentation method, an input image can be partitioned into multiple non-overlapped parts (i.e., super-pixels), with each part often having irregular form of region boundary and likely containing homogeneous visual information. 
As introduced above, the contribution of each super-pixel to the model prediction can be obtained by preserving the single super-pixel and masking the other super-pixels as the input and collecting the model output response of the predicted class to the majority-masked input. 

In practice, due to the imperfect performance of any single super-pixel segmentation method, some super-pixels may contain parts of both object region and background region, resulting in the importance map where part of background regions also have relatively higher responses. To alleviate such an issue, multiple super-pixel segmentation methods are employed, and multiple importance maps based on these segmentation methods are then averaged to estimate the contribution of each input element (e.g., pixels) to the original model prediction. 
The average importance map may be further improved by running the above process once more (i.e., second run), in which the super-pixel segmentation methods are performed on the average importance map rather than the original input image. In addition, when generating each majority-masked input, 
the highly smoothed version of the original input image is used to fill the corresponding masked regions. 

Compared to the sliding window strategy, the partitioned parts by the pre-segmentation strategy have more precise and reasonable region boundaries particularly for image data. This in turn often leads to the final importance map with clear boundaries between object regions and background regions, thus more precisely locating the image regions which contribute to the model prediction. Note that both input partition strategies also work when input is a sequence of items. For example, when input is a sentence as in the sentiment analysis task, any input can be partitioned into words or phrases with either the sliding window strategy or appropriate pre-segmentation strategy.

\subsection{Comparison with relevant studies}

The proposed PAMI framework can provide interpretation of model prediction without requiring to know model structure and parameter information. 
In contrast, most existing interpretation methods  requires either part of or the whole model details. For example, CAM and its variants need the feature maps from certain convolutional layers and part of model parameters in order to obtain the final class activation map~\cite{zhou2016learning,selvaraju2017grad,chattopadhay2018grad,wang2020score,liu2022partial}, and the gradient-based methods need all the model details to calculate gradient information over model layers~\cite{springenberg2015striving, selvaraju2017grad, chattopadhay2018grad, srinivas2019full, sundararajan2017axiomatic}. 
One exception is the occlusion method~\cite{zeiler2014visualizing,petsiuk2018rise} which does not require model details as the proposed PAMI framework. PAMI can be considered as an opposite version of the occlusion method, i.e., only preserving an input part versus only removing or occluding an input part for estimating the contribution of the input part to the model prediction. In image classification, occluding part of the foreground object in the image may not significantly affect the model prediction because the model can use the other object parts in the image for confident prediction. As a result, occlusion method may neglect contribution of certain object parts to the original model prediction, and often performs worse than the proposed PAMI.

Because the proposed PAMI framework can consider the well-trained model as a black-box, it can potentially work for various backbone structures (e.g., both CNN and Transformer backbones). In contrast, the majority of interpretation methods were proposed for the CNN backbone, and specific modifications are often required when applying existing interpretation methods (e.g., CAM or Grad-CAM) to other backbones like Transformer~\cite{chefer2021transformer,li2022transcam} and graph neural networks~\cite{pope2019explainability, barcelo2020logical, dehmamy2019understanding}.
Another merit of the proposed PAMI framework is its potential usage in multiple tasks with different input formats. While this study mainly use the image classification task for evaluation of the PAMI framework, PAMI in principle can be applied to various model prediction tasks, such as image caption and sentiment analysis, where the input data can be in the format of sentence or image. In contrast, most existing interpretation methods do not work across tasks without further modifications or extensions.

The most relevant interpretation methods are  RISE~\cite{petsiuk2018rise} and ScoreCAM~\cite{wang2020score} which also estimate the importance map based on linear combination of input masks with weights from model outputs. However, RISE is based on a large set of randomly generated masks and ScoreCAM is based on the feature maps at last convolutional layer of the (CNN) model, both leading to low-resolution and often inappropriate importance maps.


\section{Experiments}




\subsection{Experimental setup}

In this study, three image classification datasets ImageNet-2012~\cite{ILSVRC15}, Pascal VOC 2007~\cite{pascal-voc-2007}, and COCO 2014~\cite{lin2014microsoft} were mainly used for evaluation of the proposed PAMI method. 
In addition, an image caption dataset COCO~\cite{lin2014microsoft} and sentiment analysis dataset Sentiment140~\cite{Sentiment140} were also employed to show the wide applications of the proposed method. All the models were from the publicly released resources and evaluated on the corresponding validation or test set (see Table~\ref{tab:datasets} for more details). 

\begin{table}[!btp]
\centering
\renewcommand\arraystretch{1.2}
\caption{Models and datasets used in  experiments.}
\label{tab:datasets}
\resizebox{\linewidth}{!}{
\begin{tabular}{c|c|c}
\hline \hline
Task                            & Model source         & Dataset                                                                                                            \\ \hline
\multirow{5}{*}{Cassification} & VGG19bn from PyTorch~\cite{Paszke_PyTorch_An_Imperative_2019} & \begin{tabular}[c]{@{}c@{}}50000 images of ImageNet-2012~\cite{ILSVRC15} \\ validation set with 1000 classes\end{tabular}          \\ \cline{2-3} 
                                & VGG16 from TorchRay~\cite{torchray}  & \begin{tabular}[c]{@{}c@{}}Ffirst 1000 images of Pascal VOC 2007~\cite{pascal-voc-2007} \\ test set with 20 classes\end{tabular}       \\ \cline{2-3} 
                                & VGG16 from TorchRay~\cite{torchray}  & \begin{tabular}[c]{@{}c@{}}First 1000 images of COCO 2014~\cite{lin2014microsoft} \\ instances validation set with 80 classes\end{tabular} \\ \hline
Image caption  & ClipCap~\cite{mokady2021clipcap}    & COCO 2014~\cite{lin2014microsoft}                                                                                                       \\ \hline
Sentiment analysis              &    Transfomers libary~\cite{Wolf_Transformers_State-of-the-Art_Natural_2020}                 &       Sentiment140~\cite{Sentiment140} \\ 
\hline \hline
\end{tabular}

}
\end{table}


By default, for the sliding window strategy of the proposed PAMI, circular window with radius 40 pixels and step size 6 pixels was used to generate local image regions. For the pre-segmentation strategy, four super-pixel segmentation algorithms, i.e.,  felzenszwalb~\cite{felzenszwalb2004efficient}, SLIC~\cite{ren2003learning}, watershed~\cite{meyer1992color} from scikit-image library~\cite{scikit-image}, and SEEDS~\cite{bergh2012seeds} from the OpenCV  library~\cite{opencv_library} were utilized respectively for pre-segmentation of each image into multiple regions. 
Considering region of interest (i.e., relevant region to the model prediction) could vary a lot over images, each segmentation algorithm was run multiple times with different hyper-parameter settings to generate sets of local regions at different scales (see the supplementary~\ref{appendix:Hyperparameter_configuration} for detailed settings). 
The importance maps over all hyper-parameter settings and all the four pre-segmentation algorithms were averaged as the estimated importance map.
A Gaussian kernel with size $49\times49$ pixels and standard deviation 100 was used to generate the smoothed (blurred) image for region masking.

The proposed PAMI was compared with widely used visualization methods for interpretation of model predictions, including Gradient~\cite{simonyan2014deep}, GradCAM~\cite{selvaraju2017grad}, ScoreCAM~\cite{wang2020score}, RISE~\cite{petsiuk2018rise}, FullGrad~\cite{srinivas2019full}, MASK~\cite{fong2017interpretable}, Occlusion~\cite{zeiler2014visualizing}, GuidedBP~\cite{springenberg2015striving}, SmoothGrad~\cite{smilkov2017smoothgrad} and LRP ~\cite{montavon2017explaining}. The default hyper-parameter setting  for each method was adopted (see supplementary~\ref{appendix:Hyperparameter_Others} for details).
Besides qualitative evaluation, quantitative evaluation was also performed using the pointing game~\cite{zhang2018top} and the insertion metric~\cite{petsiuk2018rise}. In the pointing game, it measures whether the pixel with the highest activation in the importance map is successfully within the image region of the object corresponding to the 
interpreted class, with `hit' for success and `miss' for failure. {The average hit rate over all classes} is used to measure performance of each method. For the insertion metric, it gradually restores the original pixels in the blurred version of the original image, with pixels having higher activation in the importance map restored earlier. Higher insertion score would indicate a better performance of interpretation.

\begin{figure*}[htbp]
\centering
\includegraphics[width=1\textwidth]{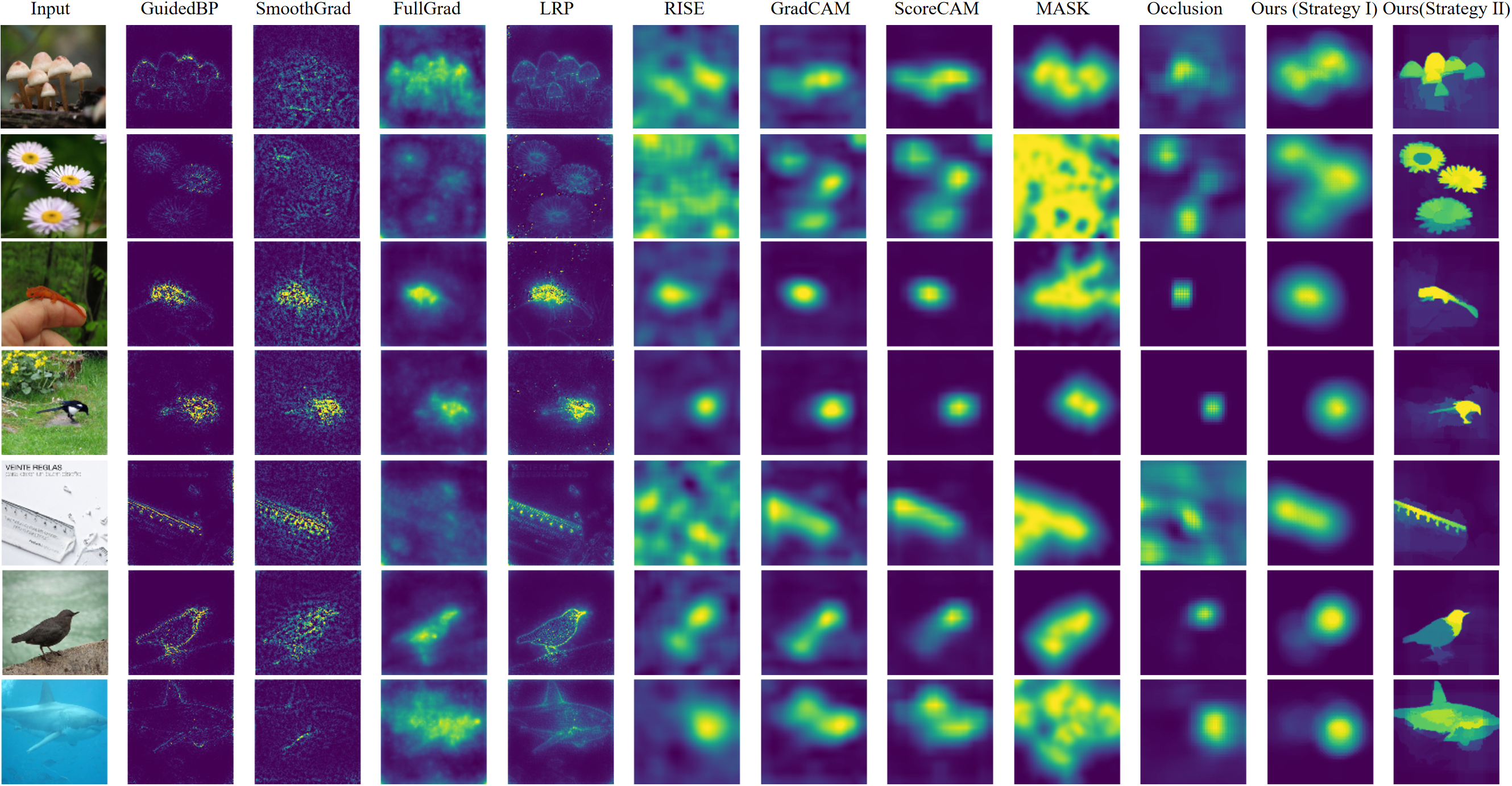}
\caption{Qualitative evaluation of the proposed PAMI method on the ImageNet-2012 dataset. The first column list the input images to the classifier. The last two columns are the importance maps from the proposed PAMI method with the two strategies respectively, and all the other columns are from the representative strong baseline methods. In each importance map or heatmap, higher activation is in yellow and lower activation is in blue.}

\label{fig:qualitative}
\end{figure*}

\subsection{Qualitative evaluation}

The efficacy of the proposed PAMI method was extensively evaluated on the ImageNet-2012 data. Figure~\ref{fig:qualitative} demonstrates the visualization results on multiple representative images with the VGG19bn model. {The visualization results were generated with respect to the model output of the ground-truth class for each image.}
It can be observed that, the proposed PAMI with the pre-segmentation strategy for input partition (last column) can often precisely and largely completely localize the object regions which are actually relevant to the specific model prediction, while existing methods roughly localize either object regions at low resolution, sparse part of object regions, disconnected pixels within object regions, or even irrelevant background regions. Multiple colors within relevant region in the importance map from the proposed PAMI method suggests that different object regions may have different degrees of contributions to the model prediction.
On the other hand, the proposed PAMI simply with the sliding window strategy can often obtain similar performance as GradCAM but without requiring model structure and parameter details.

Another observation is that the proposed PAMI method can work more stably than existing methods under challenging conditions. 
In particular, the PAMI method can well localize small-scale objects (rows 3 \& 4) and relatively large-scale objects ({row 7}) in images, and also can precisely localize the regions of multiple object instances of the same class (rows 1 \& 2). In comparison, most existing methods often perform worse  under at least some of these challenging conditions. Similar observations were also obtained on the PASCAL-VOC dataset and the COCO dataset (see supplementary~\ref{appendix:qualitative_evaluation}), consistently supporting that the proposed PAMI method is effective in providing visual evidence for interpretation of model predictions.

\subsection{Quantitative evaluation}


Although the non-existence of ground-truth or ideal interpretation for any specific model prediction makes it challenging to quantitatively evaluate any interpretation method, the pointing game~\cite{zhang2018top} and the insertion metric~\cite{petsiuk2018rise} have been proposed to roughly evaluate the performance on correctly localizing  regions relevant to model predictions. With the pointing game, Table~\ref{tab:quantitative_evaluation} (columns 2, 4, and 6) shows that the proposed PAMI method with the pre-segmentation partition strategy (last row) has similar hitting rate on the ImageNet and COCO datasets compared with the best baseline GradCAM and higher hitting rate than all the baselines on the VOC   dataset, suggesting that the local region which is considered most relevant to the model prediction by PAMI is often actually part of the object region. Similarly with the insertion metric (Table~\ref{tab:quantitative_evaluation}, columns 3, 5, and 7), {PAMI has the best performance on ImageNet and VOC datasets, and is close to the best baseline RISE on COCO dataset}, again supporting that PAMI can well localize image regions belonging to the interpreted class. More experimental details for quantitative evaluation can be found in the supplementary~\ref{appendix:details_quantitative}.





\begin{table}[tbp]
    \centering
    \tabcolsep=1mm
    \caption{Quantitative evaluation of the proposed PAMI method on the three image datasets. `Random': randomly generating a heatmap for each input image. `Center': taking the fixed image center position as the highest activation point for each input image.}
    \label{tab:quantitative_evaluation}
    \resizebox{\linewidth}{!}{
    \begin{tabular}{clcclcclcc}
\hline \hline
\multirow{2}{*}{Method}                                        & \multirow{2}{*}{} & \multicolumn{2}{c}{ImageNet} &  & \multicolumn{2}{c}{VOC}                 &  & \multicolumn{2}{c}{COCO} \\ \cline{3-4} \cline{6-7} \cline{9-10} 
                                                               &                   & Pointing     & Insertion     &  & Pointing                    & Insertion &  & Pointing   & Insertion   \\ \hline
Random  && 47.89 & - && 33.39   & -  && 11.27   & -     \\
Center  && 81.96 & - && 70.79   & -  && 25.97   & -    \\
Gradient~\cite{simonyan2014deep} && 83.14  & 0.1928 && 72.61 & 0.3321  && 34.65 & 0.1585  \\
GuidedBP~\cite{springenberg2015striving} && 83.95 & 0.2632 && 71.14 & 0.4737 && 32.83 & 0.1935  \\
Occlusion~\cite{zeiler2014visualizing}   && 84.53 & \underline{0.5741} && 84.49 & 0.6753 && 54.83  & 0.3229 \\
MASK~\cite{fong2017interpretable}  && 84.49& 0.4867 && 76.30 & 0.5616 && 49.78 & 0.2664  \\
RISE~\cite{petsiuk2018rise} && 91.58 & 0.5460 && 82.43 & \underline{0.6885} && \underline{56.95} & \bf{0.3305} \\
SmoothGrad~\cite{smilkov2017smoothgrad} && 86.51 & 0.2494 && 75.38  & 0.3824 && 39.45 & 0.1753   \\
GradCAM~\cite{selvaraju2017grad}&& \bf{93.22} & 0.5154 && \underline{87.45} & 0.5720 && \bf{57.95} & 0.2660 \\
ScoreCAM~\cite{wang2020score} && 92.01 & 0.5191 && 86.51 & 0.6030 && 55.01  & 0.2656 \\
FullGrad~\cite{srinivas2019full} && 87.01 & 0.5045 && 77.58 & 0.5049  && 44.52 & 0.2362 \\
Ours (Strategy I) && 89.17  &0.5566	 && 74.95 & 0.6133 && 48.19 & 0.2688\\
Ours (Strategy II) && \underline{92.32} & \bf{0.5965} && \bf{87.87} & \bf{0.7213} && 56.85 & \underline{0.3291}      \\ 
\hline \hline
\end{tabular}
}
\end{table}

\subsection{Generality of the proposed PAMI method}

\begin{figure*}[htbp]
\centering
\includegraphics[width=.85\linewidth]{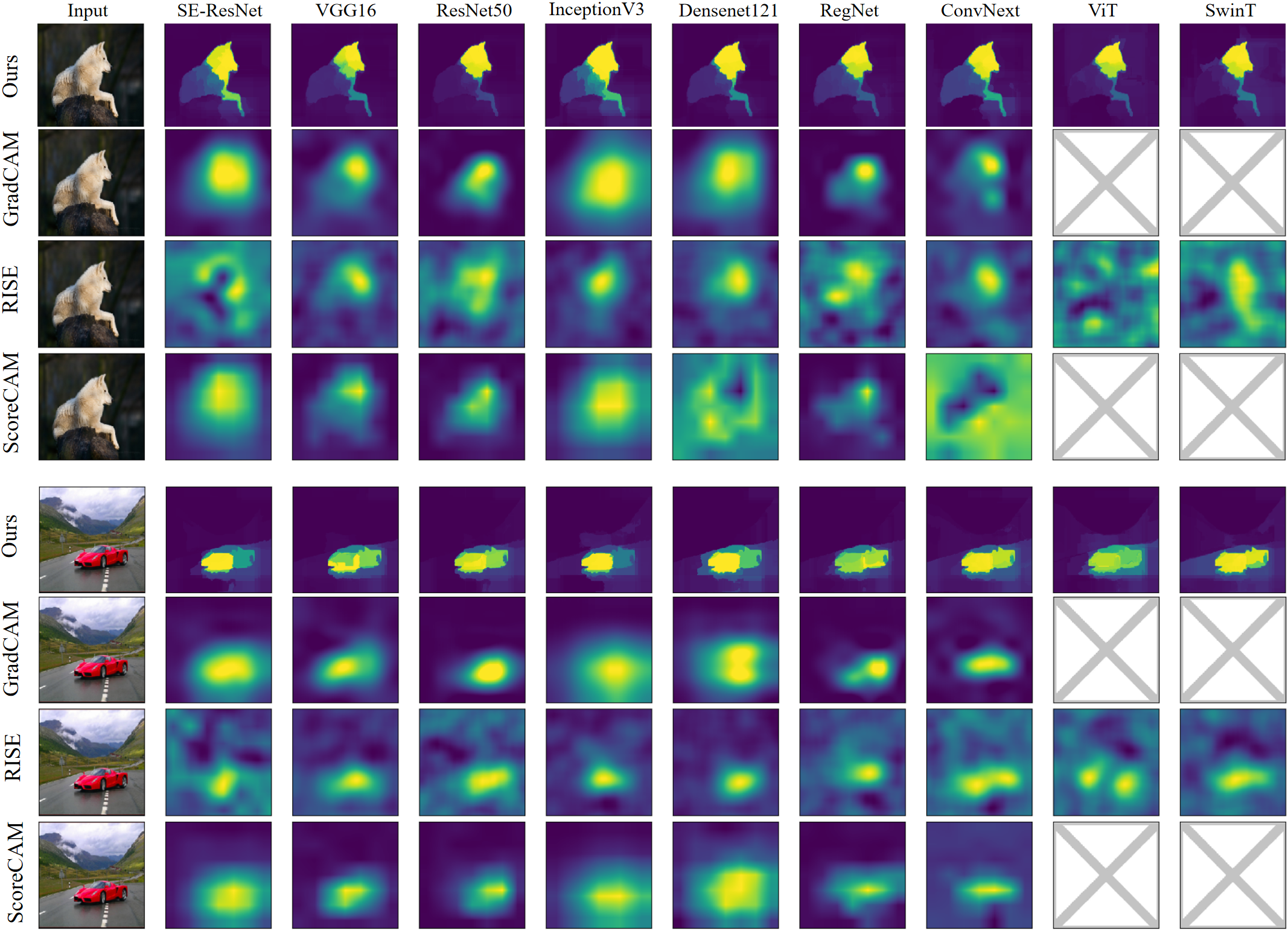}
\caption{Representative visualization results from the proposed method and representative baselines with different model backbones. Cross sign means the relevant baseline is not working on the corresponding model backbone. 
}
\label{fig:backbones}
\end{figure*}

To evaluate the generality of the proposed method, well-trained deep learning classifiers with multiple different backbones were employed, including {VGG16~\cite{simonyan2014very}, ResNet50~\cite{he2016deep}, SE-ResNet~\cite{hu2018squeeze}, InceptionV3~\cite{szegedy2016rethinking}, DenseNet121~\cite{huang2017densely}, RegNet-X-16GF~\cite{radosavovic2020designing}, ConvNext-Tiny~\cite{liu2022convnet}, ViT-L-16~\cite{dosovitskiy2020image} and SwinT-Tiny~\cite{liu2021swin}}. From Figure~\ref{fig:backbones}, we can see that the proposed PAMI method can robustly and precisely localize the object regions which are relevant to the model prediction for each input image, regardless of the classifier backbones. In contrast, for each representative baseline method, the importance maps often change over model backbones and even may not work for the Transformer backbone ViT and SwinT. 
This confirms that the proposed PAMI method is more stable and can be applied to interpretation of model predictions with various model backbones. 
Please see more results with consistent observations from the supplementary~\ref{SG}. 

\subsection{Sensitivity and ablation study}
\begin{figure}[htbp]
\includegraphics[width=1\linewidth]{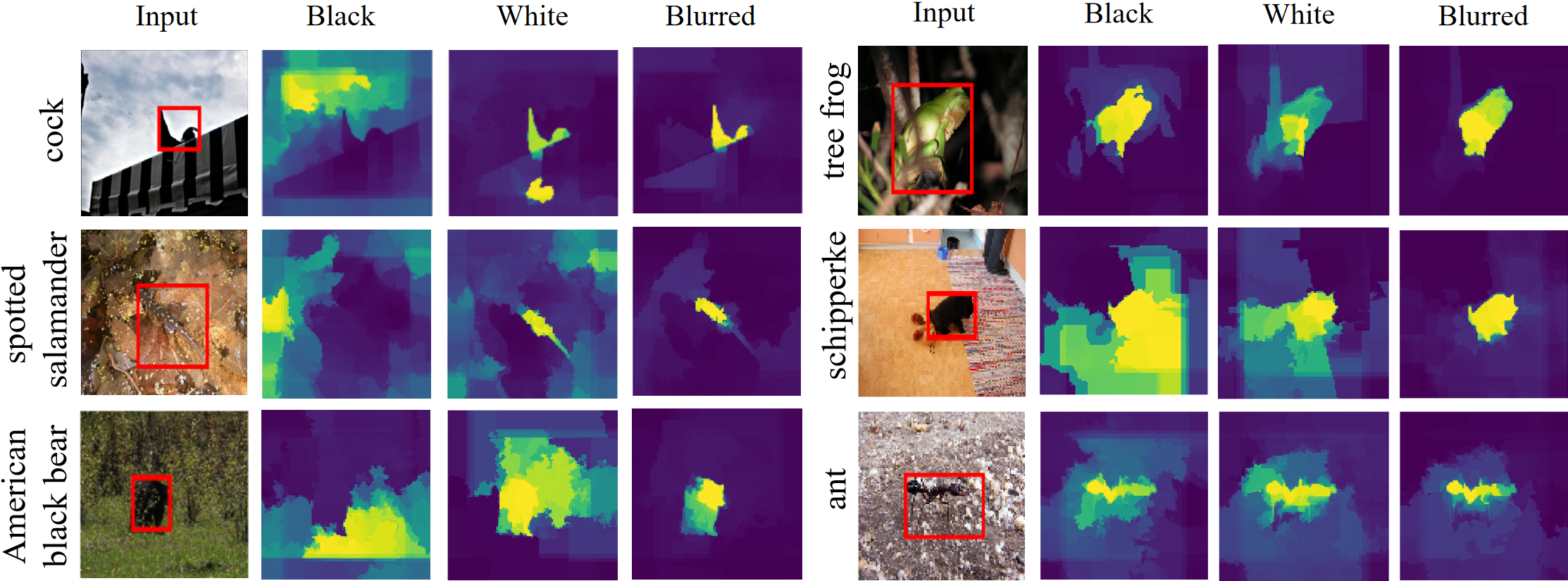}
\caption{Importance map of an representative input based on different  masking operators. `Black/White/Blurred': masking types. Red boxes in images: object regions relevant to model predictions.}

\label{fig:masking}
\end{figure}

In the proposed method, masking majority of the input is one necessary step to estimate the contribution of each single region or part of the input. There are multiple choices for the masking operator. When input is an image, the to-be-masked region could be replaced by constant intensity value such as 0 (i.e., becoming {black}) or 255 (i.e., becoming white), or by the blurred version of the input image. Figure~\ref{fig:masking} demonstrates exemplar results with different masking operators, which shows that masking by blurred region (`Blurred') results in better separation between the background regions and the object regions of interest at the first run, in turn leading to better importance maps with clearer boundaries between background and object regions at the second (i.e., final) run. 
When the majority of the input is replaced with extreme black or white pixels, the modified input image becomes further from the original class distribution in the feature space compared to the modified image with blurred region, which makes the model prediction unstable and therefore may not faithfully represent the importance of the preserved local region.

\begin{figure}[htbp]
\includegraphics[width=1\linewidth]{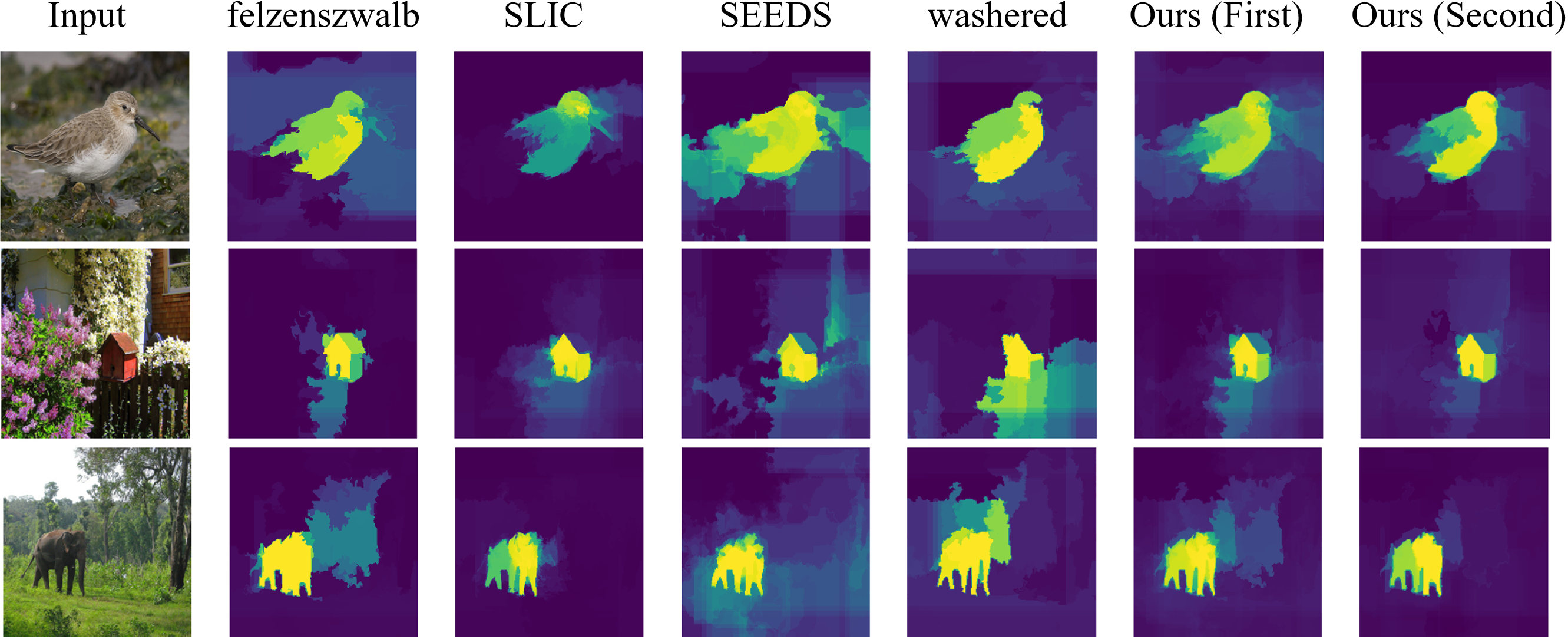}

\caption{Effect of applying multiple pre-segmentation algorithms.
From left to right: input, importance maps from {four} individual pre-segmentation algorithms, and average importance maps  at the first and the second run respectively.
}
\label{fig:average}
\end{figure}

In addition, the average of multiple importance maps from multiple pre-segmentation algorithms often results in better visualization than that of using a single pre-segmentation algorithm, as demonstrated in Figure~\ref{fig:average} ({columns 2-5 vs. column 6}). Figure~\ref{fig:average} also shows that the second run (last column) can often refine the importance map from the first run (column 6). 
This is probably because sometimes  the adopted pre-segmentation algorithms cannot well separate  background regions from object regions  at the first run, but the initially estimated importance map from the first run provides alternative information for pre-segmentation algorithms to well separate background regions from object regions.
Note that the proposed PAMI is independent of the adopted pre-segmentation algorithms, and better pre-segmentation algorithms can be adopted to replace the current ones in the future.  {Please see more ablation study results with consistent observations from the supplementary~\ref{appendix:ablation}.}



\subsection{Extensive applications of the proposed PAMI}

\begin{figure}[!htbp]
\centering
\includegraphics[width=1\linewidth]{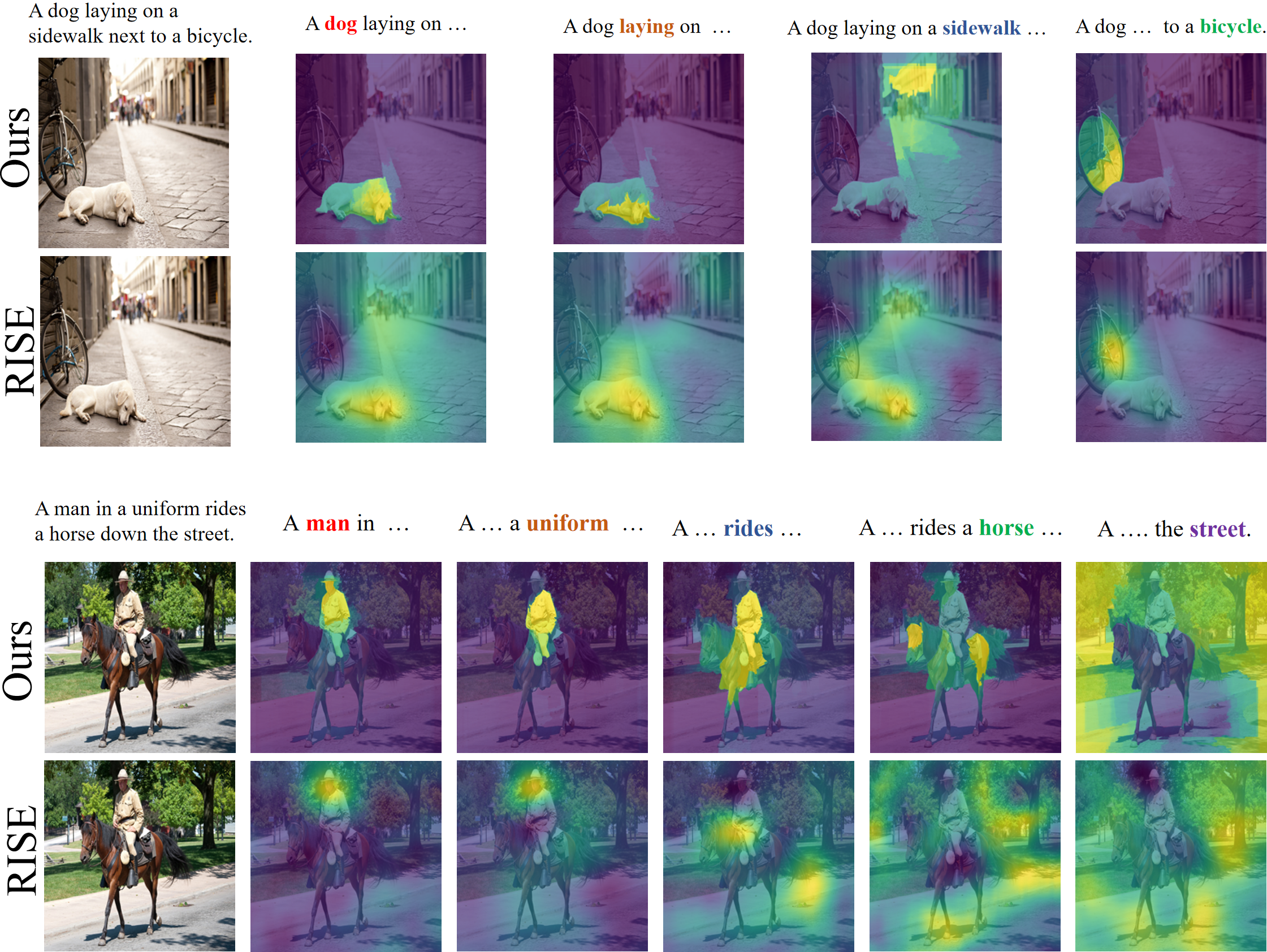}
\caption{Two exemplar visualization results from the proposed method and the strong baseline RISE for the image caption task. 
}

\label{fig:multi_task}
\end{figure}

The proposed PAMI method is expected to work for multiple types of prediction tasks. For example, based on a well-trained image caption model~\cite{mokady2021clipcap}, the PAMI method can well localize the image regions relevant to the predicted words (e.g., `dog', `laying', `sidewalk', `bicycle') which refer to any object or behaviour in the image  (Figure~\ref{fig:multi_task}, {rows 1, 3}), while the representative baseline method {RISE} often cannot precisely localize relevant regions (Figure~\ref{fig:multi_task}, {rows 2, 4}). 
Another example is for the sentiment analysis task, where the model tries to evaluate whether the viewpoint in an input sentence is positive or negative. With  an input partition strategy (see supplementary~\ref{appendix:extensive_applications})) similar to the pre-segmentation for images, the proposed PAMI method can directly and correctly estimate the contribution of each word to the final model prediction (Figure~\ref{fig:Sentiment}). These results (also see more results in the supplementary~\ref{appendix:extensive_applications}) confirm that the proposed PAMI method can work for various tasks with different input modalities.

\begin{figure}[!htbp]
\centering
\includegraphics[width=1\linewidth]{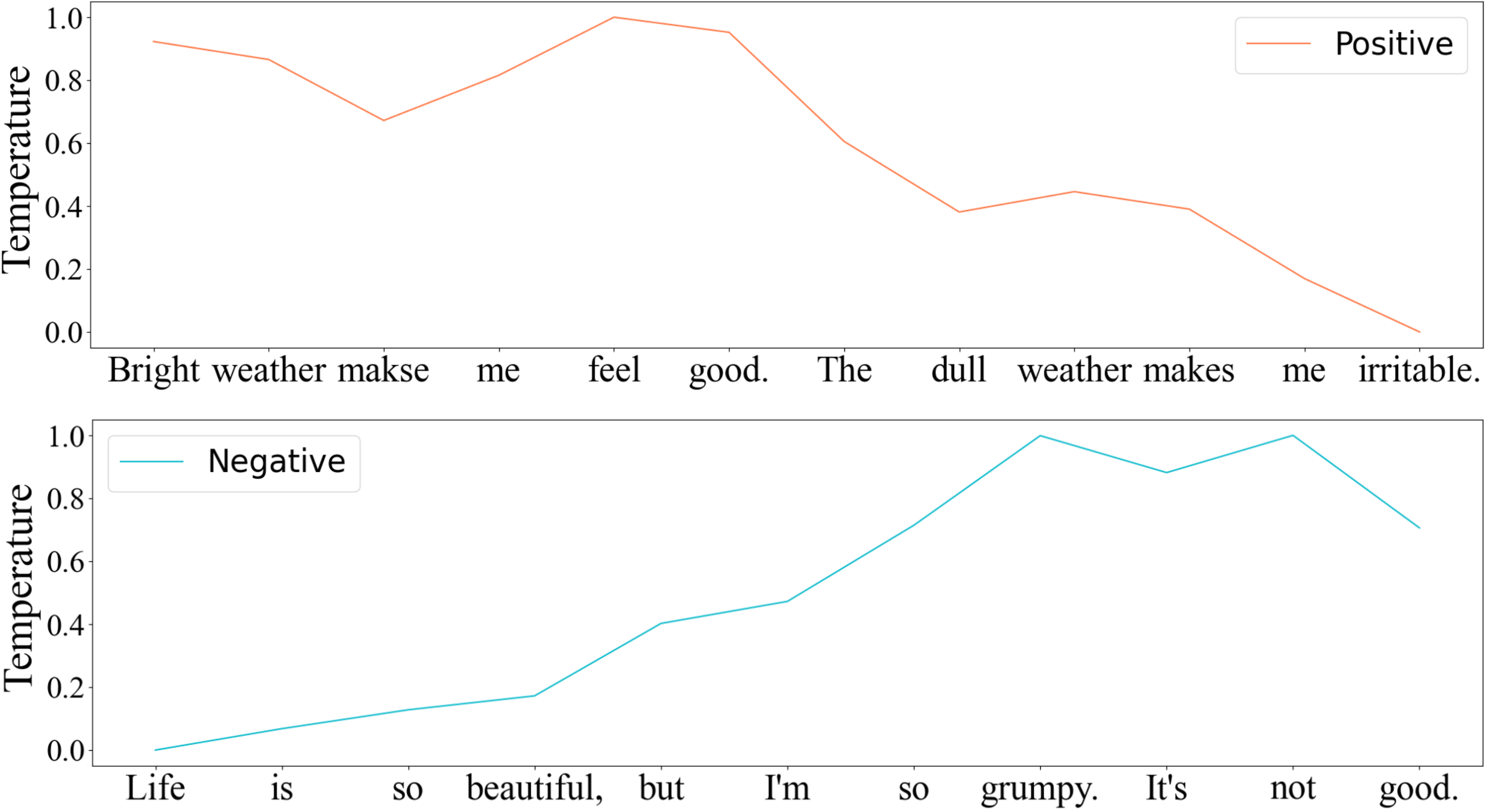}
\caption{{Two exemplar visualization results from the proposed method for the sentiment analysis task. The first row shows the contribution of each word to a positive emotion prediction, and the second row for a negative emotion prediction.}}
\label{fig:Sentiment}
\end{figure}




\section{Conclusion}
In this study, we propose a novel visualization method PAMI for interpretation of model predictions. PAMI does not requires any model parameter details and works stably across model backbones and input formats. Compared to existing visualization approaches, PAMI can more likely and precisely find the possible local input regions which contribute to the specific model prediction to some degree. 
It can be used as a plug-in component  and applied to multiple types of prediction tasks, which has been partly confirmed by  image classification, image caption, and sentiment analysis tasks. Its utility in more tasks including various natural language processing tasks will be evaluated in future work.

{\small
\bibliographystyle{ieee_fullname.bst}
\bibliography{egbib.bib}
}

\clearpage
\appendix

\section{Hyper-parameters for pre-segmentation}
\label{appendix:Hyperparameter_configuration}
The hyper-parameters of the four pre-segmentation algorithms used in both the first and the second run are summarized in  table~\ref{fig:Hyperparameter_ours}.
The hyper-parameter names used in the methods correspond to those in the skimage and cv2 packages. 

\begin{table}[h]
    \centering
    \caption{Hyper-parameter configuration for each segmentation algorithm. 
    }
    \label{fig:Hyperparameter_ours}
    \resizebox{1\linewidth}{!}{
    \begin{tabular}{c|c}
    \hline
    \hline
        Method & Hyper-parameter\\
    \hline
         felzenszwalb~\cite{felzenszwalb2004efficient} &  \makecell[c]{scale=250, 200, 150, 100, 70, 50\\ sigma=0.8\\ min\_size=784}\\ \hline
         SLIC~\cite{ren2003learning} & \makecell[c]{n\_segments=10, 20, 30, 40, 50, 60, 70, 80\\ compactness=20}\\ \hline
         SEEDS~\cite{bergh2012seeds} & \makecell[c]{num\_superpixels=10, 20,30 \\ num\_levels=5 \\ n\_iter=10} \\ \hline
         watershed~\cite{meyer1992color} & \makecell[c]{markers=10, 20, 30 \\ compactness=0.0001}
    \\ \hline \hline
    \end{tabular}
    }
\end{table}

\section{Hyper-parameters in baseline methods}
\label{appendix:Hyperparameter_Others}

Details of each baseline method and reference source code were provided in Table~\ref{tab:Hyperparameter_others}. For gradient-based methods, following the related work~\cite{simonyan2014deep}, the maximum importance value among the three channels at each spatial position was used as the final importance value for the spatial position in  the importance map. For LRP~\cite{montavon2017explaining},the values from three channels were averaged as the final result for each spatial position.

\begin{table}[h]
    \centering
    \caption{Hyper-parameter configuration for each baseline method.
    }
    \label{tab:Hyperparameter_others}
    \resizebox{\linewidth}{!}{
    \begin{tabular}{c|c|c}
    \hline
    \hline
        Method & Hyper-parameter & Code Source\\
    \hline
         GradCAM~\cite{felzenszwalb2004efficient} & The last layer of feature extractor & PyTorch CNN Visualizations\cite{uozbulak_pytorch_vis_2022}\\ \hline
         ScoreCAM~\cite{wang2020score} & The last layer of feature extractor & TorchCam\cite{torcham2020}\\ \hline
         Occlusion~\cite{zeiler2014visualizing} & \makecell[c]{strides=6, shapes=(3, 40, 40)} & Captum~\cite{kokhlikyan2020captum}\\ \hline
         RISE~\cite{petsiuk2018rise} & \makecell[c]{num\_mask=4000, cell\_size=7 \\ probability=0.5} & TorchVex~\cite{torchvex}\\ \hline
         MASK~\cite{fong2017interpretable} & \makecell[c]{TV\_beta=3, lr=0.1, max\_iterations=500 \\ l1\_coeff=0.01, tv\_coeff=0.2} & TorchVex~\cite{torchvex} \\ \hline
         FullGrad~\cite{bergh2012seeds} & - & original implementation\cite{srinivas2019full}
    \\ \hline \hline
    \end{tabular}
    }
\end{table}

\section{More qualitative evaluation}
\label{appendix:qualitative_evaluation}

\subsection{More results on ImageNet-2012}
\begin{figure*}[htbp]
\centering
\includegraphics[width=1.\linewidth]{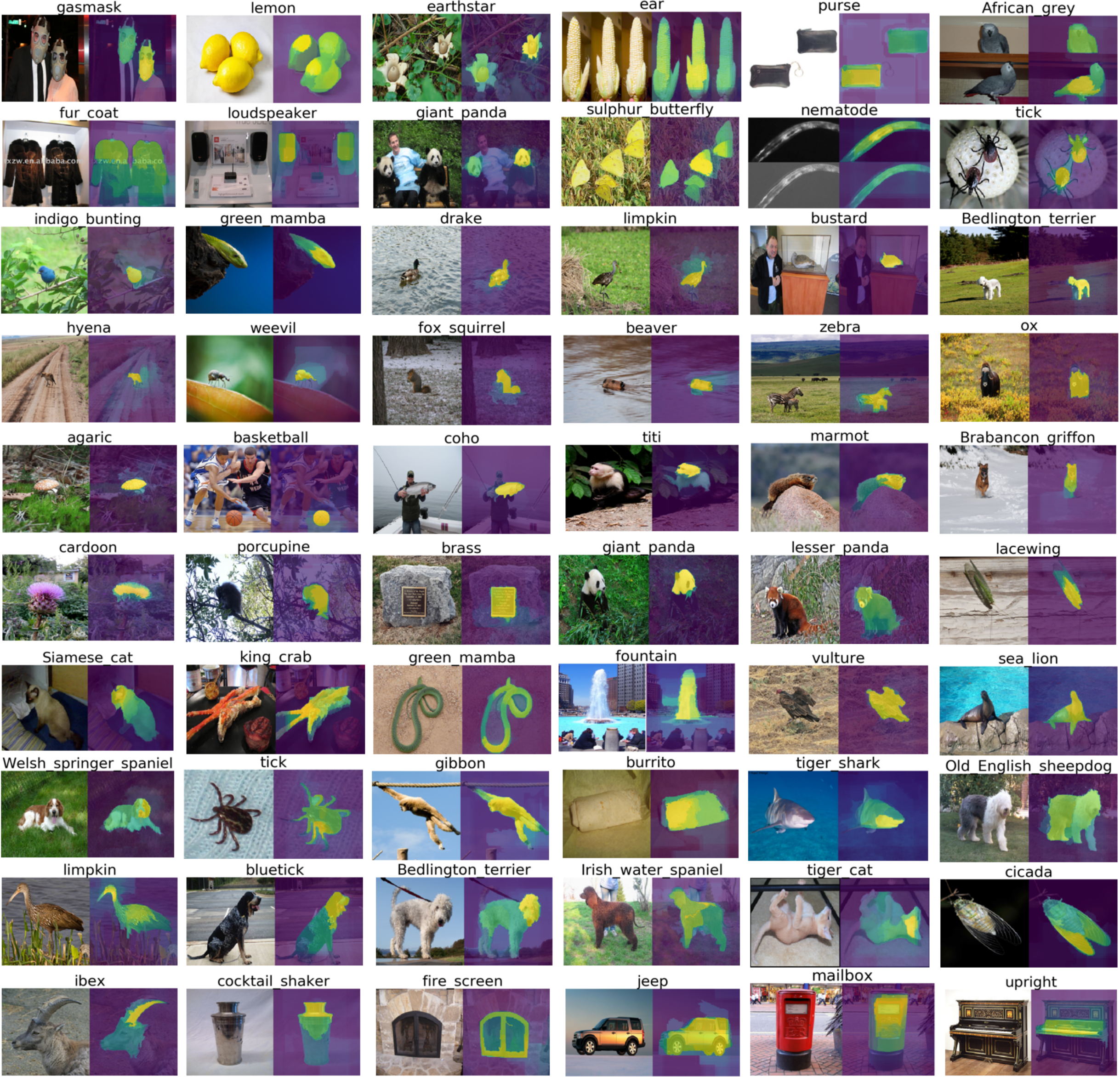}
\caption{More qualitative evaluation of the proposed PAMI method on the ImageNet-2012 dataset. For each pair: input image is on the left and the visualization result from the proposed PAMI is on the right.}
\label{fig:appendix_imagenet}
\end{figure*}

More qualitative evaluation of the proposed PAMI method on the ImageNet-2012 dataset~\cite{ILSVRC15} can be seen in Figure~\ref{fig:appendix_imagenet}, supporting the effectiveness of the method.

\subsection{More results on Pascal VOC 2007}
\begin{figure*}[htbp]
\centering
\includegraphics[width=1.\linewidth]{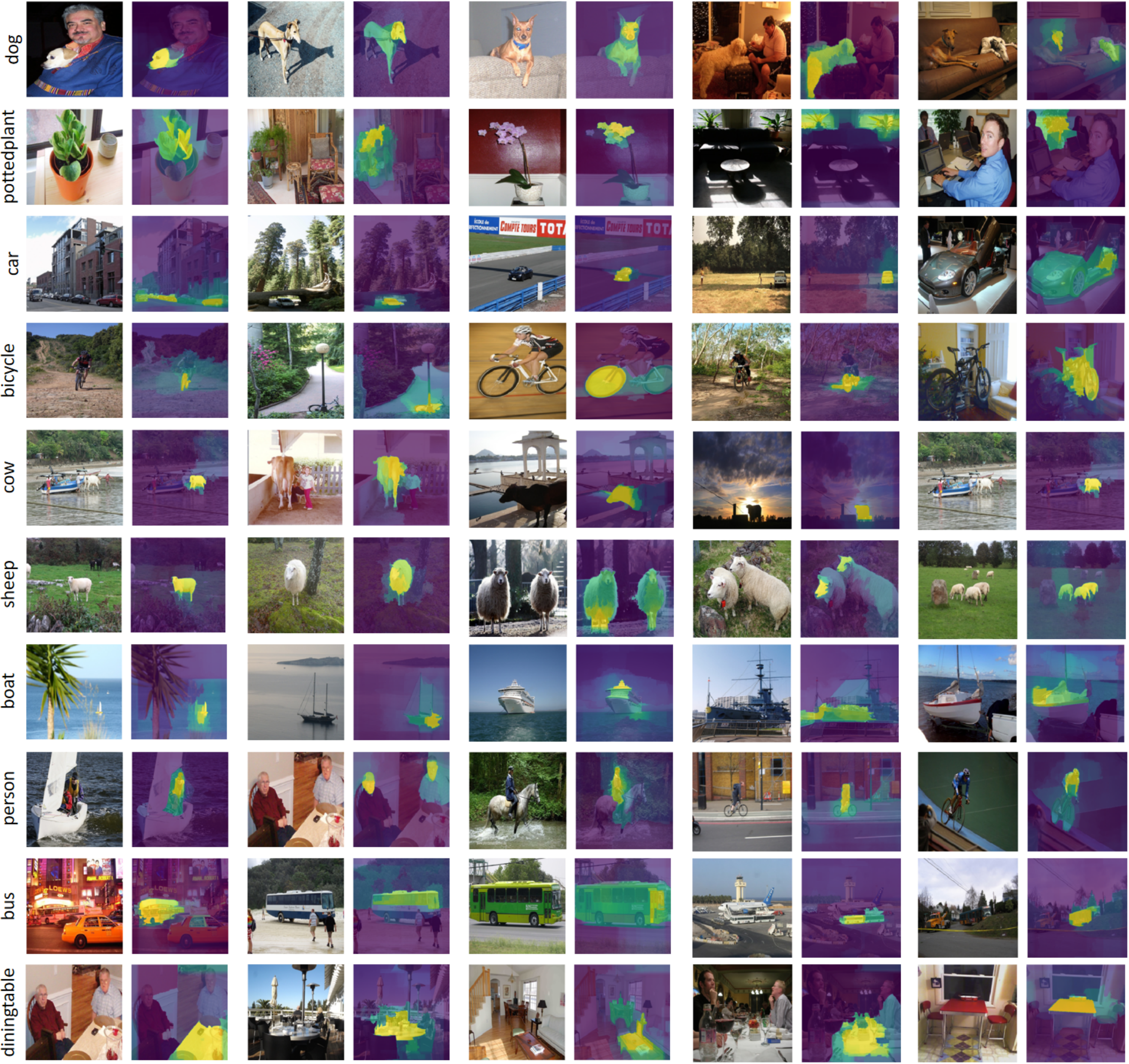}
\caption{Qualitative evaluation of the proposed PAMI method on Pascal VOC 2007 dataset. Note that each input was resized to be square for demonstration.}
\label{fig:appendix_pascal}
\end{figure*}

The effectiveness of the proposed PAMI method was also validated 
on the Pascal VOC 2007 dataset~\cite{everingham2009pascal}, as shown in Figure~\ref{fig:appendix_pascal}. 

\subsection{More results on COCO 2014}
\begin{figure*}[htbp]
\centering
\includegraphics[width=1.\linewidth]{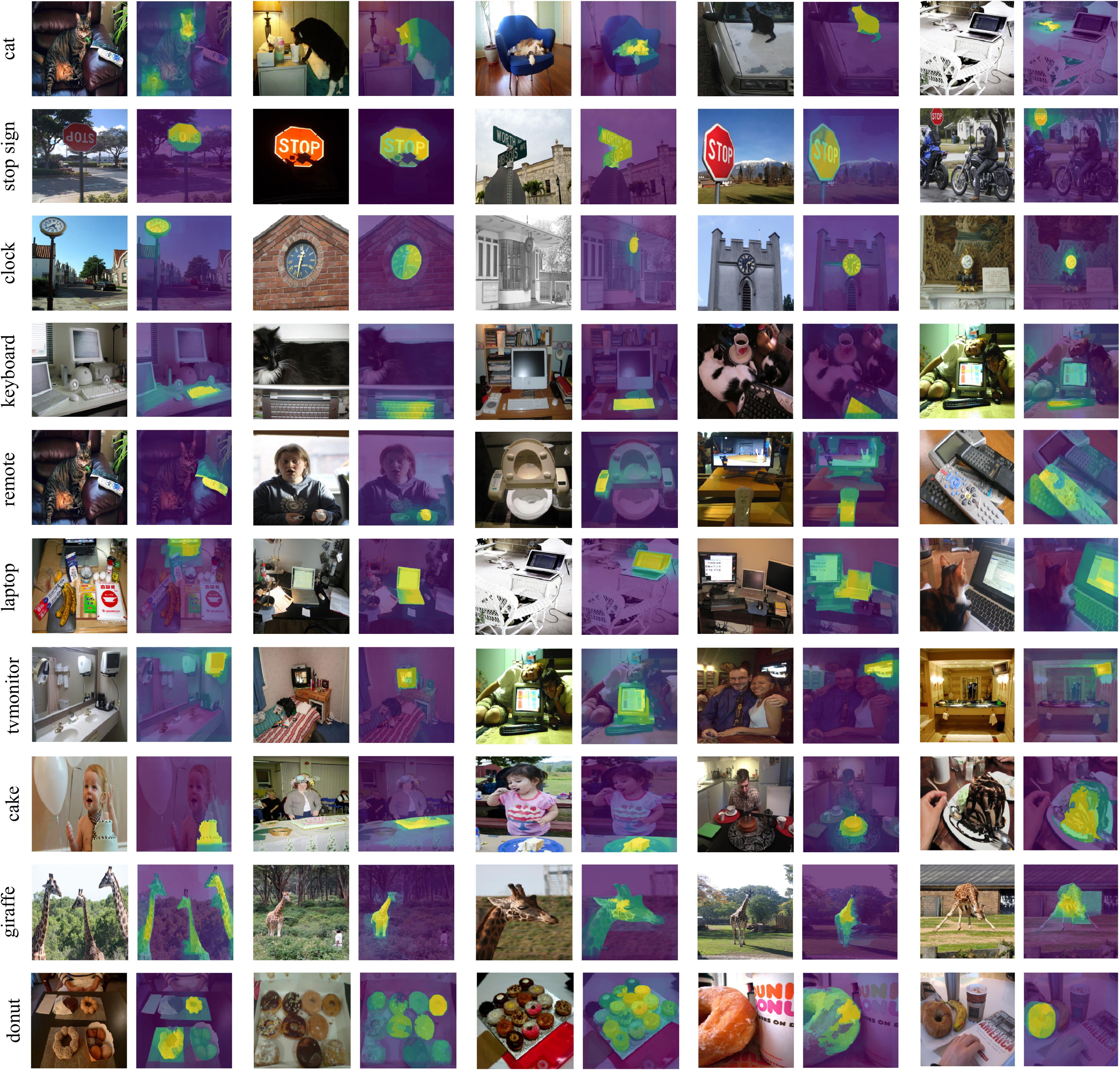}
\caption{Qualitative evaluation of the proposed PAMI method on COCO 2014 dataset.  Note that each input was resized to be square for demonstration.}
\label{fig:appendix_coco}
\end{figure*}

The effectiveness of the proposed PAMI method was also validated 
on the COCO 2014 dataset~\cite{lin2014microsoft}, as shown in Figure~\ref{fig:appendix_coco}. 

\section{Details of quantitative evaluation}
\label{appendix:details_quantitative}
\begin{figure}[htbp]
\centering
\includegraphics[width=1.\linewidth]{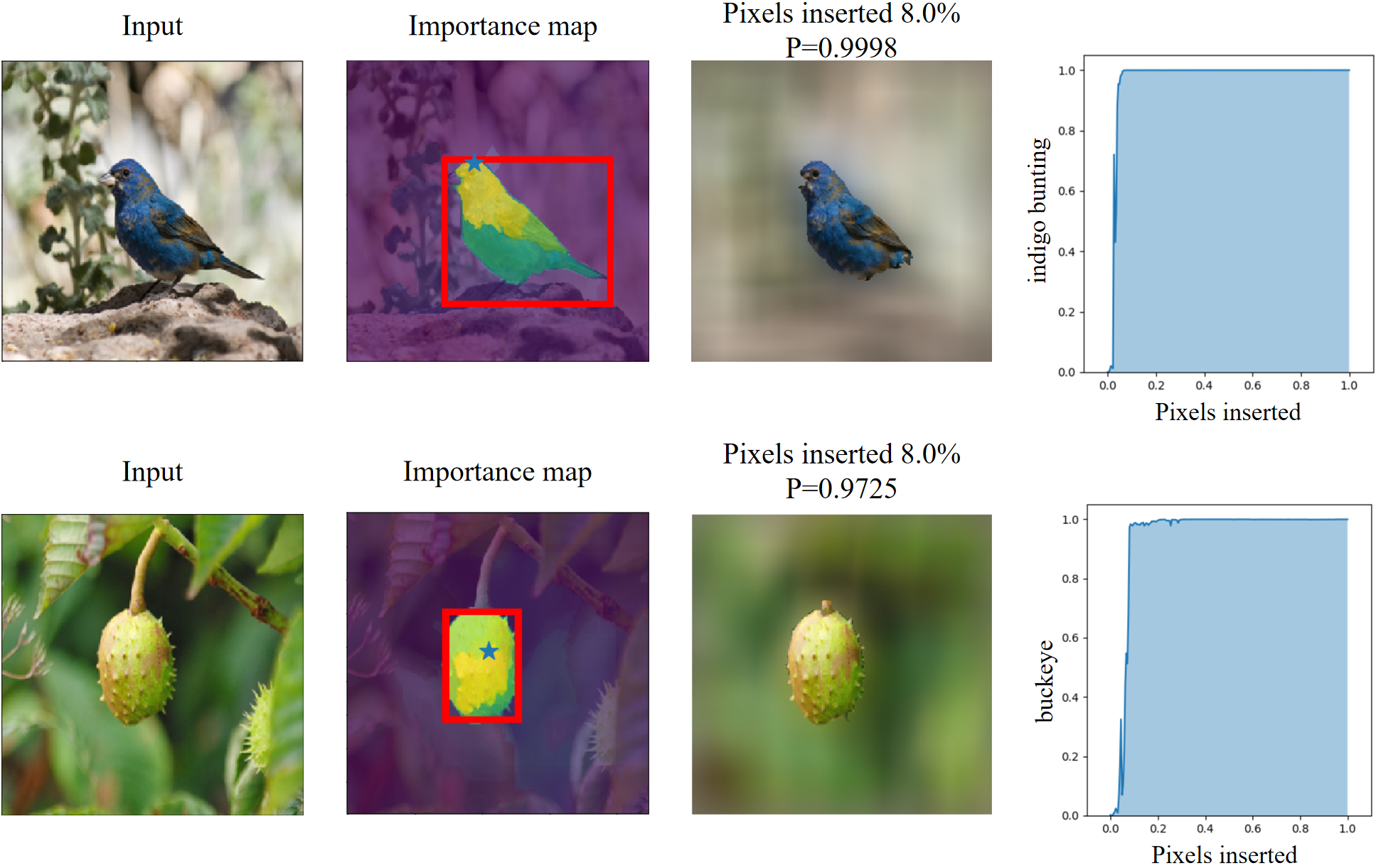}

\caption{Two examples of quantitative evaluation. First column:  original input image. Second column: the importance map and the most important pixel marked with asterisks. Third column: image after restoring a certain percentage of pixels and the prediction probability of the image being the ground-truth class by the model. Fourth column: the insertion curve and the area under the curve as the insertion score.}
\label{appendix:quantitative_evaluation}
\end{figure}

For the pointing game, the process provided by TorchRay~\cite{torchray} was followed. On the Pascal VOC and the COCO datasets, the evaluation code provided by TorchRay was directly used, and on the ImageNet dataset, the same process was performed by marking points that fall within the bounding box of the object as hits and the rest as misses. For the insertion metric, a Gaussian kernel with size $49 \times 49$ pixels and standard deviation $100$ was used to generate the blurred image where we gradually restore the original pixels from. The importance map for quantitative evaluation were generated with respect to the model output of the ground-truth class for each image. Two exemplar results were shown in Figure~\ref{appendix:quantitative_evaluation}.


\section{Generality of the PAMI}
\label{SG}

\begin{figure}[htbp]
\centering
\includegraphics[width=1.\linewidth]{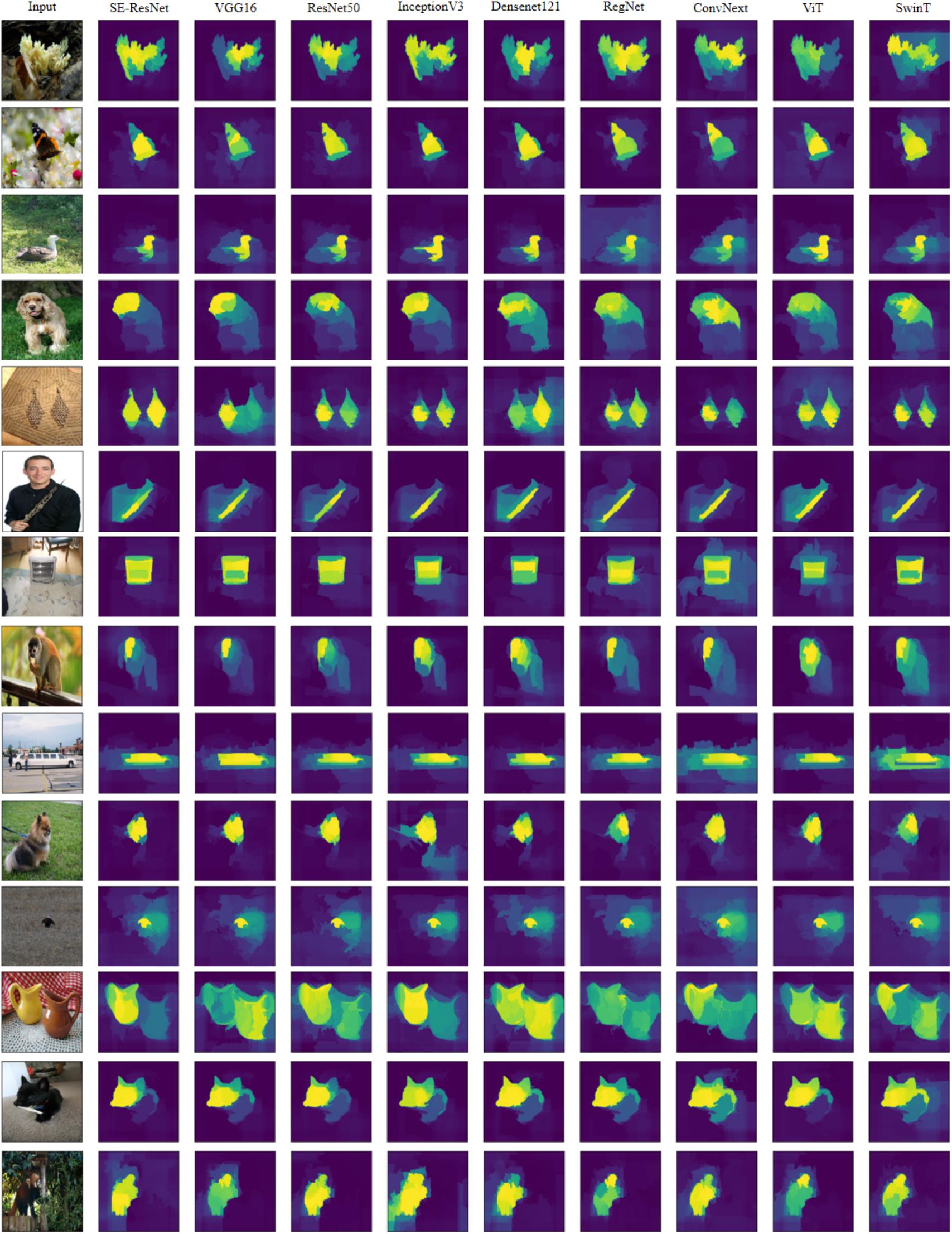}

\caption{More representative visualization results from the proposed method with different model backbones.}
\label{appendix:generality}
\end{figure}

More results with different model backbones were shown in Figure~\ref{appendix:generality}, supporting the generality of the proposed method. 

\section{More ablation study results}
\label{appendix:ablation}
\subsection{Effect of different masking operators}
\begin{figure}[htbp]
\centering
\includegraphics[width=1\linewidth]{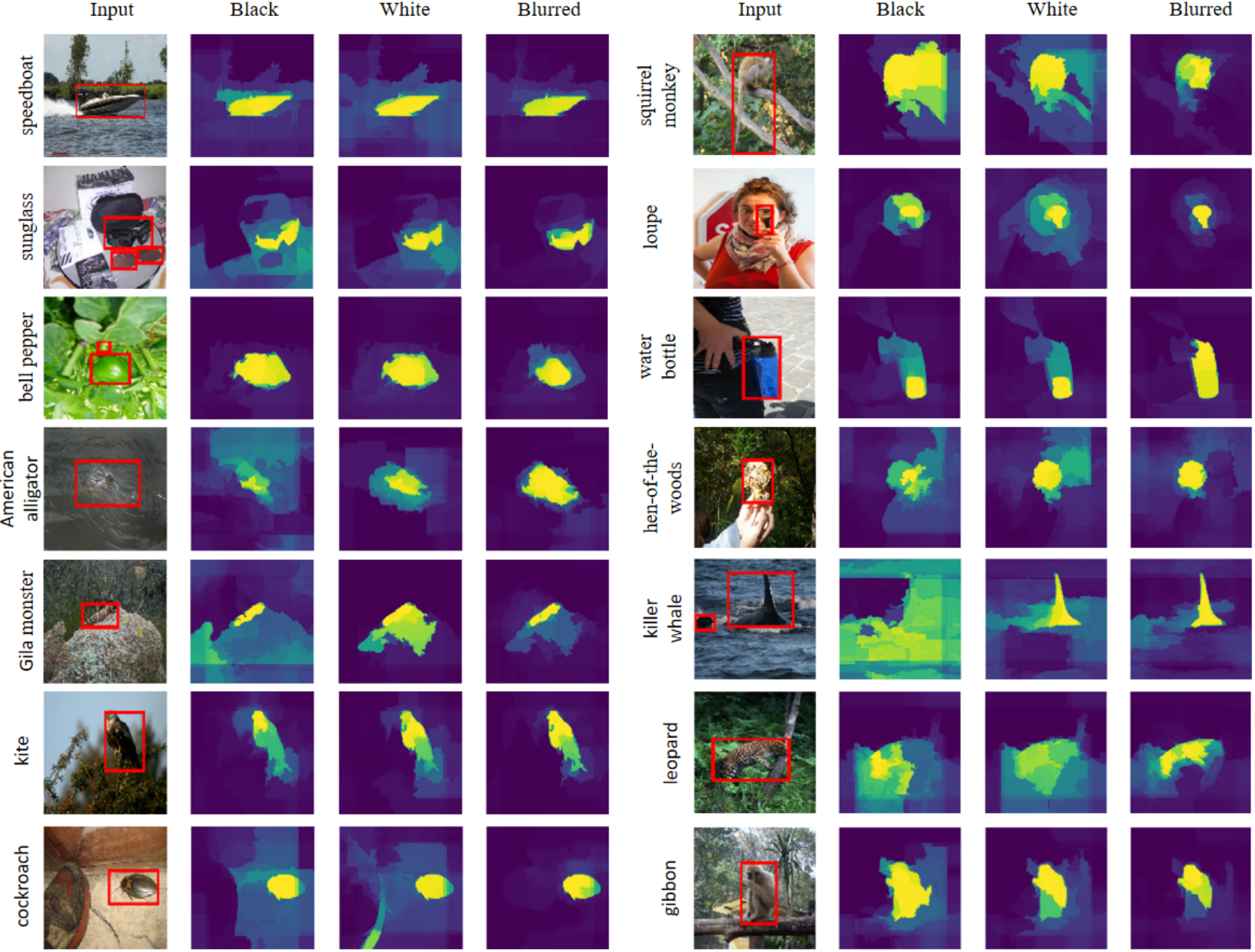}
\caption{Importance maps of more representative inputs based on different masking operators.}
\label{sup:blured}
\end{figure}

More visualization results based on different masking operators were provided in Figure~\ref{sup:blured}, which shows that the blurred version results in better results especially when image background is complex.

\subsection{Effect of multiple pre-segmentation algorithms}
\begin{figure}[htbp]
\centering
\includegraphics[width=1\linewidth]{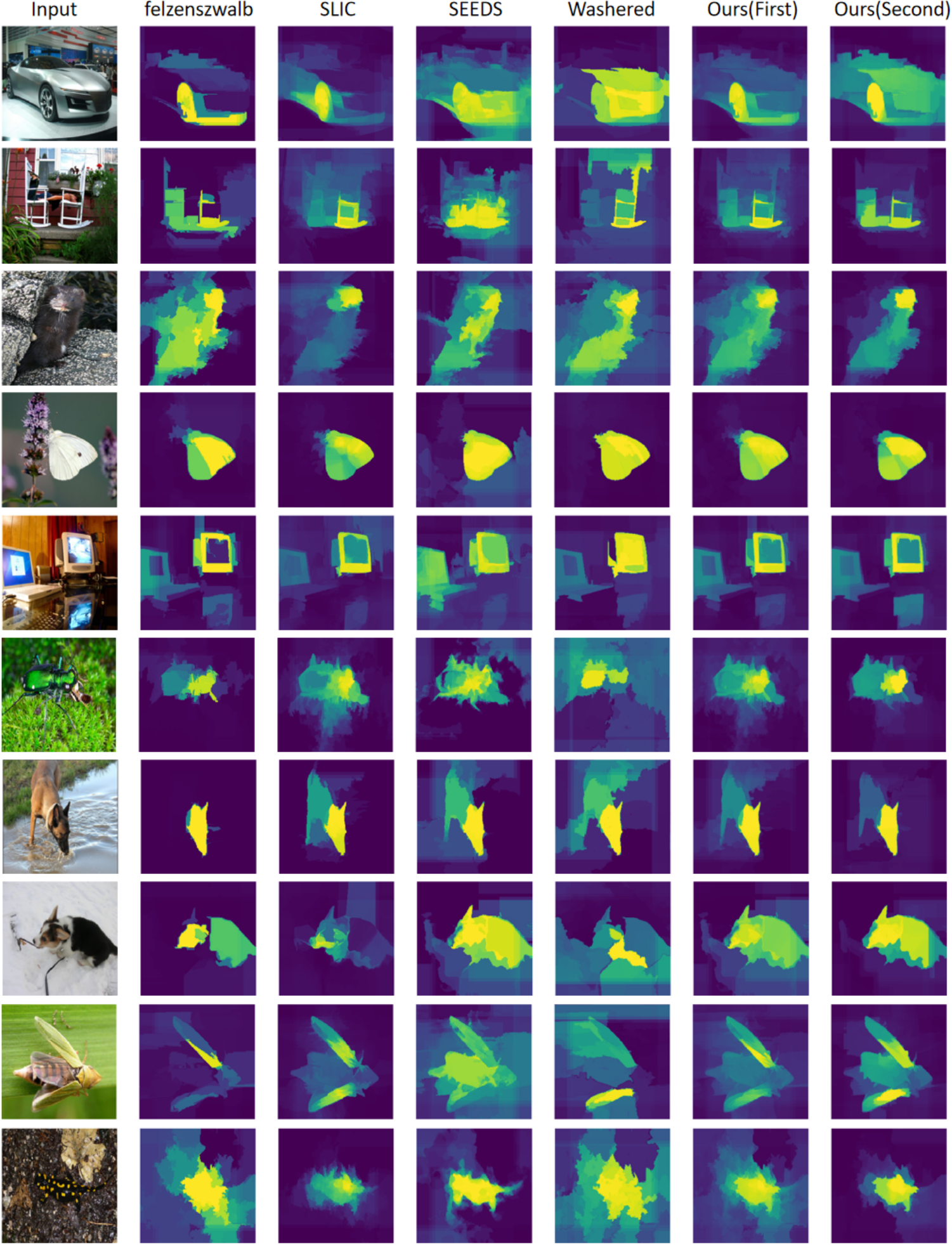}
\caption{Effect of applying multiple pre-segmentation algorithms.
From left to right: input, importance maps from four individual
pre-segmentation algorithms, and average importance maps at the
first and the second run respectively.
}
\label{sup:multi_segmentation}
\end{figure}

More results in Figure~\ref{sup:multi_segmentation} shows that using multiple pre-segmentations and two runs result in better visualizations.

\section{Extensive applications of PAMI}
\label{appendix:extensive_applications}

\subsection{Image caption task}
\begin{figure*}[htbp]
\begin{center}
\includegraphics[width=1\linewidth]{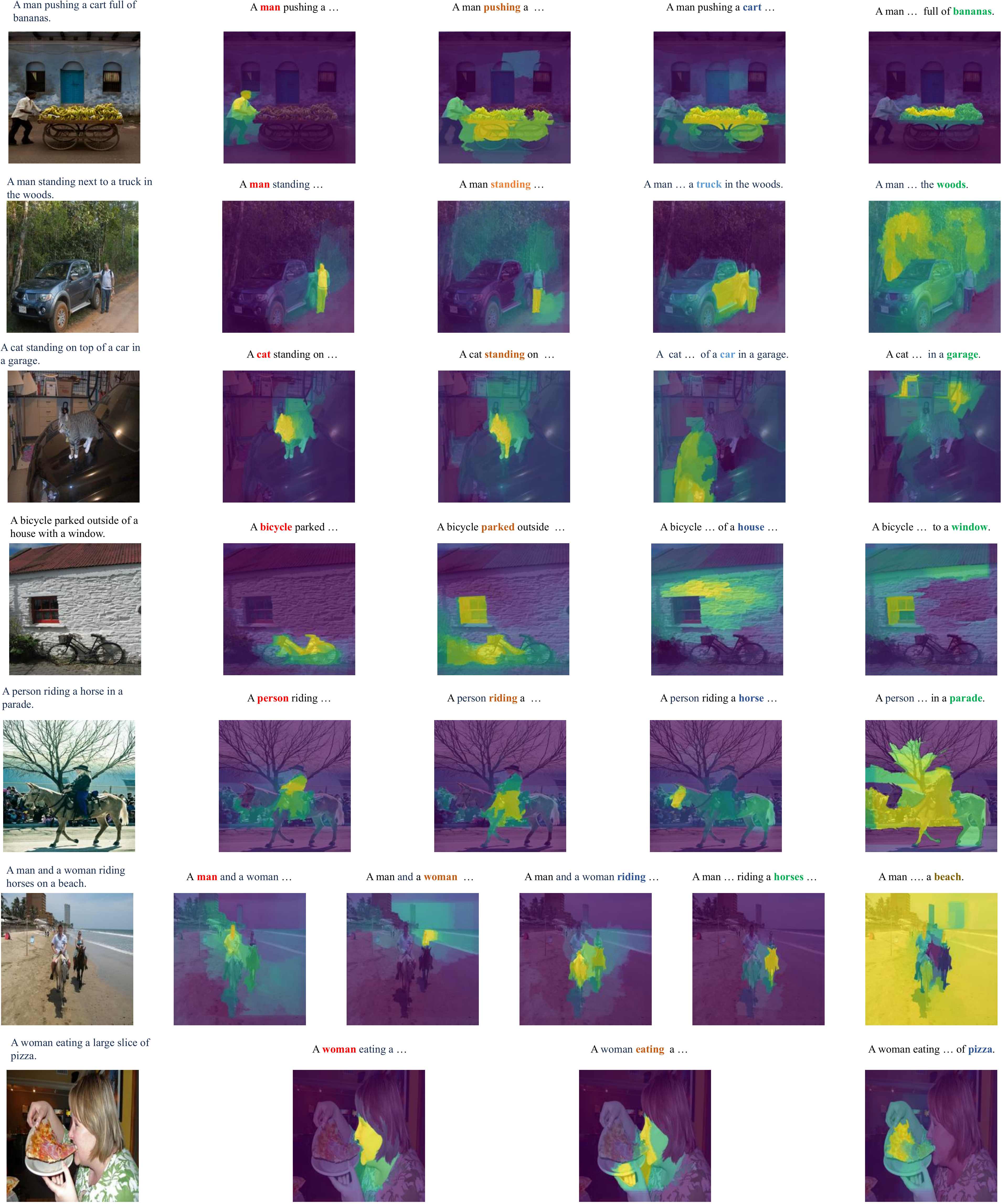}
\end{center}
\caption{More visualization results from the proposed
method for the image caption task.}
\label{fig:appendix_imagecaption}
\end{figure*}

More experimental results for the image caption task on the COCO dataset can be seen in Figure~\ref{fig:appendix_imagecaption}.

\subsection{Sentiment analysis task}
\begin{figure*}[htbp]
\begin{center}
\includegraphics[width=1.\linewidth]{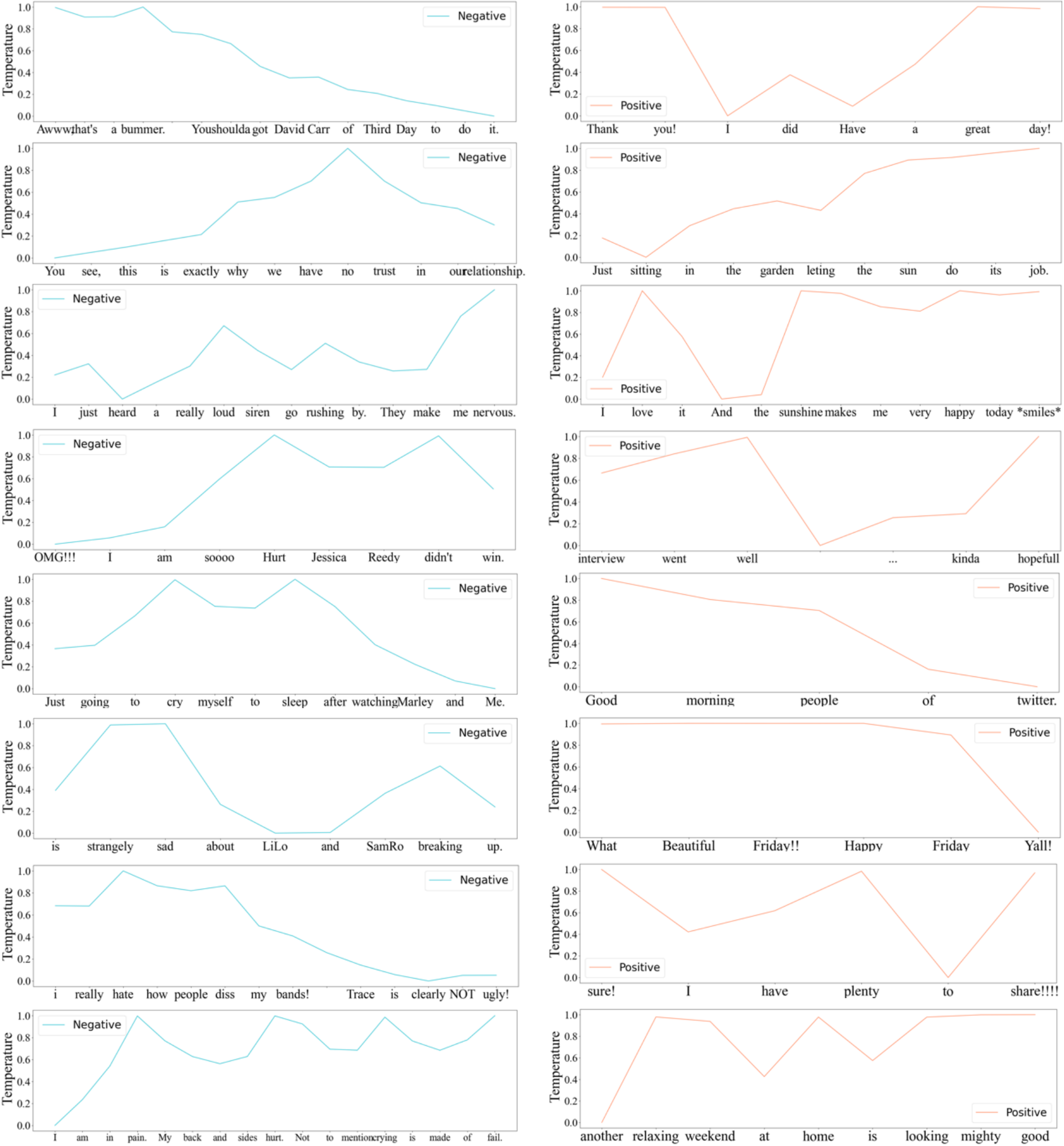}
\end{center}
\caption{More visualization results from the proposed
method for the sentiment analysis task.}
\label{fig:appendix_sentiment}
\end{figure*}

More test sentences in Sentiment140~\cite{Sentiment140} were randomly selected for effectiveness evaluation of the proposed method in sentiment analysis tasks. The experimental results can be seen in Figure~\ref{fig:appendix_sentiment}.

\end{document}